\documentclass[english,american]{IEEEtran}
\usepackage[T1]{fontenc}
\usepackage[utf8]{inputenc}
\usepackage{xcolor}
\usepackage{array}
\usepackage{bm}
\usepackage{multirow}
\usepackage{amsmath}
\usepackage{amssymb}
\usepackage{graphicx}
\PassOptionsToPackage{normalem}{ulem}
\usepackage{ulem}

\makeatletter

\providecommand{\tabularnewline}{\\}
\providecolor{lyxadded}{rgb}{0,0,1}
\providecolor{lyxdeleted}{rgb}{1,0,0}

\DeclareRobustCommand{\lyxsout}[1]{\ifx\\#1\else\sout{#1}\fi}

\usepackage{color}
\usepackage{psfrag}\usepackage{cite}\usepackage{subfigure}

\usepackage{flushend}

\usepackage[top=0.75in, bottom=1.05in, left=0.625in, right=0.625in]{geometry}

\author{

\IEEEauthorblockN{Zheng~Xing and Junting~Chen}

\IEEEauthorblockA{School of Science and Engineering (SSE) and Shenzhen Future Network of Intelligence Institute (FNii-Shenzhen) \\ The Chinese University of Hong Kong, Shenzhen, Guangdong 518172, China}
\thanks{The work was supported in part by Guangdong S\&T Programme with Grant No. 2024B0101030002, the Basic Research Project No. HZQB-KCZYZ-2021067 of Hetao Shenzhen-HK S\&T Cooperation Zone, the National Natural Science Foundation of China No. 62171398, the Shenzhen Science and Technology Program No. JCYJ20220530143804010 and No. KJZD20230923115104009, the Guangdong Basic and Applied Basic Research Foundation 2024A1515011206 and Research Projects No. 2017ZT07X152 and No. 2019CX01X104, the Shenzhen Outstanding Talents Training Fund 202002, the Guangdong Provincial Key Laboratory of Future Networks of Intelligence (Grant No. 2022B1212010001), and the Shenzhen Key Laboratory of Big Data and Artificial Intelligence (Grant No. ZDSYS201707251409055).}
}

\usepackage[acronym]{glossaries}
\newcommand{\newac}{\newacronym}
\newcommand{\ac}{\gls}

\newcommand{\acpl}{\glspl}

\makeglossaries

\usepackage{enumitem}

\usepackage{amsmath}   
\usepackage{cuted}

\newac{ips}{IPS}{indoor positioning system}
\newac{wlan}{WLAN}{wireless local area network}
\newac{ap}{AP}{access point}
\newac{sc}{SC}{subspace clustering}

\newac{uwb}{UWB}{ultrawideband}
\newac{hmm}{HMM}{hidden Markov model}
\newac{ema}{EM}{expectation maximization}
\newac{lbs}{LBS}{location-based service}
\newac{gpc}{GPC}{Gaussian process classification}

\newac{gpca}{GPCA}{generalized PCA}
\newac{msl}{MSL}{multistage learning}
\newac{mppca}{MPPCA}{mixture of probabilistic PCA}
\newac{alc}{ALC}{agglomerative lossy compression}
\newac{ransac}{RANSAC}{random sample consensus}
\newac{fa}{FA}{factorization-based affinity}
\newac{gpcaa}{GPCAA}{gpca-based affinity}
\newac{slbf}{SLBF}{spectral local best-fit flats}
\newac{lsa}{LSA}{local subspace affinity}
\newac{llmc}{LLMC}{locally linear manifold clustering}
\newac{ssc}{SSC}{sparse subspace clustering}
\newac{lrr}{LRR}{low-rank representation}
\newac{scc}{SCC}{spectral curvature clustering}
\newac{mfb}{MFB}{matrix factorization-based}
\newac{mssc}{MSSC}{multi-view low-rank sparse subspace clustering}
\newac{dsc}{DSC}{deep subspace clustering}
\newac{sssc}{SSSC}{Stochastic sparse subspace clustering}
\newac{mdssc}{MDSSC}{}
\newac{nmi}{NMI}{normalized mutual information}
\newac{svm}{SVM}{support vector machine}
\newac{tdoa}{TDOA}{time difference of arrival}
\newac{gmm}{GMM}{Gaussian mixture model}
\newac{pca}{PCA}{Principal Component Analysis}
\newac{dnn}{DNN}{deep neural network}
\newac{mlp}{MLP}{multilayer perceptron}
\newac{admm}{ADMM}{alternating direction method of multipliers}
\newac{gan}{GAN}{generative adversarial networks}
\newac{rsrp}{RSRP}{reference signal received power}
\newac{ss}{SS}{synchronization signal}
\newac{fim}{FIM}{Fisher information matrix}
\newac{crlb}{CRLB}{Cramer-Rao Lower Bound}
\newac{1nn}{1-NN}{1 Nearest-Neighbor}
\newac{dft}{DFT}{discrete Fourier transform}
\newac{evd}{EVD}{Eigen Value Decomposition}
\newac{ssb}{SSB}{secondary synchronization block}
\newac{lstm}{LSTM}{long short-term memory}
\newac{ppp}{PPP}{Poisson Point Process}
\newac{ofdm}{OFDM}{Orthogonal Frequency-Division Multiplexing}
\newac{csi}{CSI}{channel state information}
\newac{mimo}{MIMO}{multiple-input multiple-output}
\newac{nlos}{NLOS}{non-line-of-sight}
\newac{snr}{SNR}{signal-to-noise ratio}
\newac{bs}{BS}{base station}
\newac{rssi}{RSSI}{received signal strength indicator}
\newac{aoa}{AoA}{angle of arrival}
\newac{mle}{MLE}{maximum likelihood estimation}
\newac{mse}{MSE}{mean squared error}
\newac{wrt}{w.r.t.}{with respect to}
\newac{uav}{UAV}{unmanned aerial vehicle}
\newac{rss}{RSS}{received signal strength}
\newac{sinr}{SINR}{signal-to-interference-and-noise ratio}
\newac{los}{LOS}{line-of-sight}
\newac{pdf}{PDF}{probability density function}
\newac{gps}{GPS}{global positioning system}
\newac{mae}{MAE}{mean absolute error}
\newac{ula}{ULA}{uniform linear arrays}
\newac{knn}{KNN}{$k$-nearest neighbors}

\usepackage{babel}

\usepackage{array}

\usepackage{varwidth}

\addto\captionsamerican{}
\addto\captionsamerican{}

\addto\captionsenglish{}
\addto\captionsenglish{}

\makeatother

\usepackage{babel}
\begin{document}
\title{Unsupervised Radio Map Construction in Mixed LoS/NLoS Indoor Environments}
\maketitle
\begin{abstract}
Radio maps are essential for enhancing wireless communications and
localization. However, existing methods for constructing radio maps
typically require costly calibration processes to collect location-labeled
\ac{csi} datasets. This paper aims to recover the data collection
trajectory directly from the channel propagation sequence, eliminating
the need for location calibration. The key idea is to employ a \ac{hmm}-based
framework to conditionally model the channel propagation matrix, while
simultaneously modeling the location correlation in the trajectory.
The primary challenges involve modeling the complex relationship between
channel propagation in \ac{mimo} networks and geographical locations,
and addressing both \ac{los} and \ac{nlos} indoor conditions. In
this paper, we propose an \ac{hmm}-based framework that jointly characterizes
the conditional propagation model and the evolution of the user trajectory.
Specifically, the channel propagation in \ac{mimo} networks is modeled
separately in terms of power, delay, and angle, with distinct models
for LOS and NLOS conditions. The user trajectory is modeled using
a Gaussian-Markov model. The parameters for channel propagation, the
mobility model, and LOS/NLOS classification are optimized simultaneously.
Experimental validation using simulated MIMO-\ac{ofdm} networks with a
multi-antenna \ac{ula} configuration demonstrates
that the proposed method achieves an average localization accuracy
of 0.65 meters in an indoor environment, covering both LOS and NLOS
regions. Moreover, the constructed radio map enables localization with a reduced error compared to conventional supervised methods, such as \ac{knn}, \ac{svm}, and \ac{dnn}.

\end{abstract}

\begin{IEEEkeywords}
Radio map, trajectory recovery, LOS/NLOS classification, MIMO-OFDM
networks
\end{IEEEkeywords}

\section{Introduction}
\thispagestyle{empty}
The construction and utilization of radio maps have become a central
focus in wireless communications and localization research \cite{Liu:J23,Shre:J22,WanHon:J22,XinChe:C22}.
A persistent challenge in radio map construction lies in the collection
of location-labeled \ac{csi} data \cite{ZheChe:J25}, which requires significant resources
for drive testing and manual calibration. Radio maps establish crucial
links between physical locations and channel characteristics, enabling
innovative approaches for \ac{csi} acquisition, tracking, and prediction,
as well as low-latency \ac{mimo} communications and environment-aware
beamforming in mmWave massive MIMO systems \cite{WuZen:J23,XinChe:C25}.

Traditional methods for radio map construction have proposed various
strategies to reduce calibration efforts \cite{Xing:C22,XinChe:C24}. For instance, \cite{ChiMas:J22}
employed Kriging and covariance tapering techniques to construct radio
maps in massive MIMO systems using a minimal amount of location-labeled
\ac{csi} data. Similarly, \cite{WanZhu:J24} used sparse sampling
and Bayesian learning inference to construct radio maps with limited
location-labeled \ac{csi} data. However, these methods still necessitate
calibration efforts to obtain even small amounts of location-labeled
data. While some GPS-based methods \cite{Whi:J22} can provide location
information, their performance is suboptimal in indoor environments
and faces privacy access issues. 

Channel charting emerges as an alternative paradigm, utilizing \ac{csi}
embedding techniques to map high-dimensional channel data to low-dimensional
latent spaces. Building on dimensionality reduction methods, subsequent
work has explored various approaches such as autoencoders \cite{Agos:J20}
and Sammon’s mapping \cite{FerRau:J21}. However, these methods still
require a small amount of location labels, often rotating 2D or 3D
data in the latent space based on limited location labels. Additionally,
the latent space does not necessarily represent the physical world
and does not account for dense multipath propagation in NLOS/LOS cross-environments \cite{Xing:C25-2}.
Our prior work \cite{XingChen:J23ar} employed a splitting-merging
algorithm to recover trajectories but was limited to achieving only
region-level accuracy.

In this paper, we propose a \ac{hmm}-based framework that jointly
characterizes the conditional propagation model and the evolution
of user trajectories in \ac{mimo} networks. The model distinguishes
between \ac{los} and \ac{nlos} conditions, capturing the distinct
channel propagation dynamics of each. The channel propagation in MIMO
networks is separately modeled in terms of key parameters including
power, delay, and angle, with distinct models for both LOS and NLOS
environments. The mobile user’s trajectory is modeled using a Gaussian-Markov
model to capture the temporal-spatial relationship between each location
and its preceding positions. The paper employs alternating optimization
to jointly optimize the parameters for channel propagation, mobility
models, and the classification of LOS/NLOS conditions.

The proposed method is validated through ray-tracing simulations conducted
in MIMO-OFDM networks, covering both \ac{los} and \ac{nlos} indoor
environments. The results demonstrate that our method achieves an
average localization error of 0.65 meters, significantly outperforming
traditional angle- and power-based localization techniques, and also
surpassing the performance of channel charting methods. Furthermore,
by leveraging the constructed radio map, the method consistently achieves
an average localization error below 1 meters, slightly outperforming
conventional supervised approaches such as DNN, KNN, and SVM. 

\begin{figure}[t]
\centering{}\includegraphics[width=1\columnwidth]{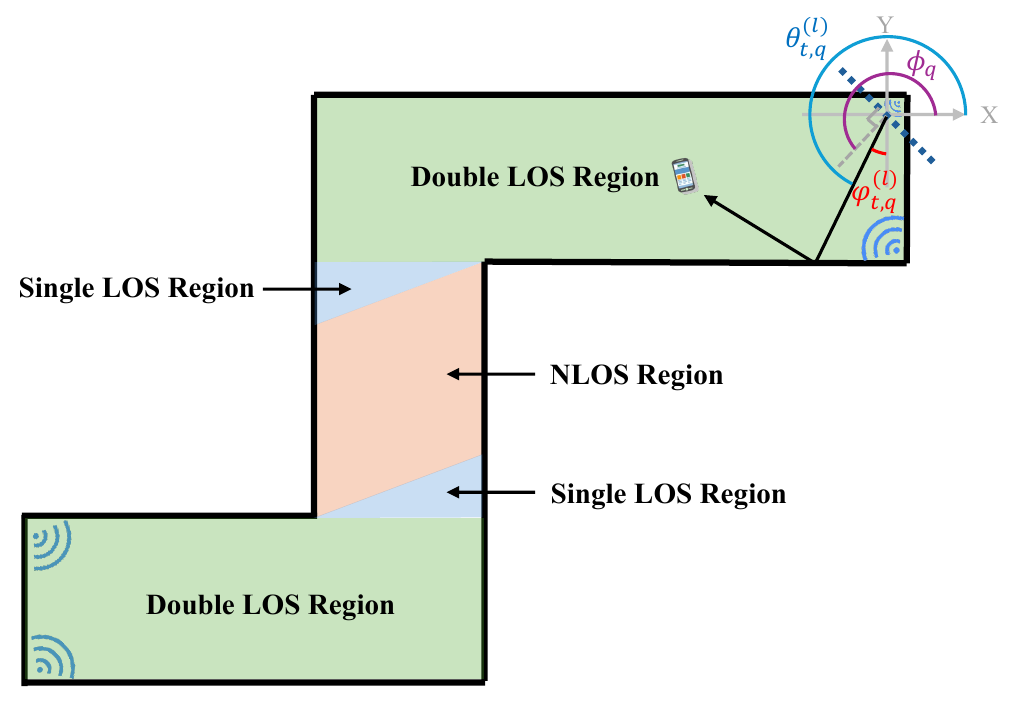}\caption{Schematic diagram of multipath signal emission angles and ULA antenna
orientation. \label{fig:antenna}}
\end{figure}

\section{System Model}

\subsection{Channel Model in MIMO-OFDM Networks}

We consider a mobile user navigating an indoor environment with static
obstructions, served by a network of $Q$ \acpl{ap} located
at known positions $\mathbf{o}_{1},\mathbf{o}_{2},\ldots,\mathbf{o}_{Q}\in\mathbb{R}^{2}$.
Each AP is equipped with a \ac{ula} consisting of
$N_{\mathrm{t}}$ transmit antennas. The orientation angle $\phi_{q}\in[0,2\pi)$
denotes the angle of the $q$-th AP’s antenna array normal vector
relative to the global coordinate system, as shown in Fig. \ref{fig:antenna}.
The system operates under a MIMO-OFDM framework with orthogonal subcarrier
allocation.

The user trajectory is represented by a discrete sequence $\{\mathbf{x}_{t}\}_{t=1}^{T}\subset\mathbb{R}^{2}$
of $T$ planar positions sampled at uniform time intervals. In the
presence of multipath propagation, the frequency-selective channel
between the $q$-th AP and the user at position $\mathbf{x}_{t}$
consists of $L$ resolvable paths. Each $\ell$-th path ($\ell=1,\ldots,L$)
is characterized by three parameters: the complex gain $\kappa_{t,q}^{(\ell)}\in\mathbb{C}$,
the propagation delay $\tau_{t,q}^{(\ell)}\in\mathbb{R}_{+}$, and
the angle of departure (AoD) $\theta_{t,q}^{(\ell)}\in[0,2\pi)$.
The effective AoD in the $q$-th AP’s local coordinate system is given
by $\varphi_{t,q}^{(\ell)}=\theta_{t,q}^{(\ell)}-\phi_{q}$, where
$\varphi_{t,q}^{(\ell)}\in(-\pi/2,\pi/2)$ due to the spatial aliasing
limits of the ULA.

The normalized steering vector for the $q$-th AP's ULA at AoD $\theta$
is defined as: 
\begin{equation}
\mathbf{a}_{q}(\theta)=\left[1,e^{-j\frac{2\pi}{\lambda}\Delta\sin\theta},\ldots,e^{-j\frac{2\pi}{\lambda}(N_{\mathrm{t}}-1)\Delta\sin\theta}\right]^{T},\label{eq:steering}
\end{equation}
where $\Delta$ denotes the inter-element spacing, $\lambda=c/f_{c}$
is the wavelength at the carrier frequency $f_{c}$, and $c=3\times10^{8}\ \mathrm{m/s}$
is the speed of light. The phase progression across array elements
follows from the plane wave assumption in far-field radiation.

The channel for the $q$-th AP on the $m$-th subcarrier is given
by $\mathbf{h}_{t,q}^{(m)}=\sum_{\ell=1}^{L}\kappa_{t,q}^{(\ell)}\cdot e^{-j2\pi\frac{m}{M}B\tau_{t,q}^{(\ell)}}\cdot\mathbf{a}(\varphi_{t,q}^{(\ell)})$,
where $B$ is the total bandwidth, $M$ is the total number of subcarriers,
and we assume $M>N_{\mathrm{t}}$. $\varphi_{t,q}^{(\ell)}$ is the
AoD relative to the antenna array’s orientation $\phi_{q}$, and $\mathbf{a}(\varphi_{t,q}^{(\ell)})$
is the steering vector of the ULA for the angle $\varphi_{t,q}^{(\ell)}$.
The overall CSI is then formed by stacking the responses from each
antenna $\mathbf{H}_{t,q}=[\mathbf{h}_{t,q}^{(1)},\mathbf{h}_{t,q}^{(2)},\dots,\mathbf{h}_{t,q}^{(M)}]\in\mathbb{C}^{N_{\mathrm{t}}\times M}$.
Here, the CSI $\mathbf{H}_{t,q}$ captures the frequency-selective
multipath effects, delay spread, and spatiotemporal dynamics characteristic
of wideband MIMO-OFDM channels.

\subsection{Power-Angle-Delay Profile in LOS and NLOS Regions}

\label{subsec:PDFmodel}

The propagation environment is classified into two distinct regions:
the \ac{los} region $\mathcal{D}_{0}\subset\mathbb{R}^{4}$ and the
\ac{nlos} region $\mathcal{D}_{1}\subset\mathbb{R}^{4}$. These regions
are defined for each AP-user pair $(\mathbf{o},\mathbf{x})$, where
$\mathbf{o}$ represents the location of the AP and $\mathbf{x}$
represents the user location. The regions satisfy the following condition:
\begin{equation}
\mathcal{D}_{0}\cup\mathcal{D}_{1}=\mathbb{R}^{4},\quad\mathcal{D}_{0}\cap\mathcal{D}_{1}=\emptyset.
\end{equation}

The radio map for each AP-user pair $(\mathbf{o},\mathbf{x})$ is
modeled as a piecewise function that distinguishes between LOS and
NLOS propagation: 
\begin{equation}
f(\mathbf{o},\mathbf{x};\mathcal{D}_{0},\mathcal{D}_{1})=\begin{cases}
f_{0}(\mathbf{o},\mathbf{x}), & \text{if }(\mathbf{o},\mathbf{x})\in\mathcal{D}_{0},\\
f_{1}(\mathbf{o},\mathbf{x}), & \text{if }(\mathbf{o},\mathbf{x})\in\mathcal{D}_{1},
\end{cases}
\end{equation}
where $f_{0}(\mathbf{o},\mathbf{x})$ and $f_{1}(\mathbf{o},\mathbf{x})$
represent the propagation models for LOS and NLOS, respectively.

The radio map model decomposes the propagation environment into two
complementary components: the geometric structure $\{\mathcal{D}_{0},\mathcal{D}_{1}\}$
that characterizes the LOS and NLOS spatial domains, and the stochastic
propagation model $\{f_{0},f_{1}\}$ that governs signal attenuation
and multipath phenomena. The geometric component $\mathcal{D}_{k}\subset\mathbb{R}^{4}$
for $k\in\{0,1\}$ encodes obstacle-induced visibility constraints
through spatial occupancy patterns of walls, furniture, and architectural
features. 

\subsubsection{Power Model with LOS/NLOS Discrimination}

The received signal strength (RSS) is estimated by the spatial-frequency
domain power aggregation:
\begin{equation}
s_{t,q}=10\log_{10}\|\mathbf{H}_{t,q}\|_{\mathrm{F}}^{2},\label{eq:H-s}
\end{equation}
where the CSI $\mathbf{H}_{t,q}\in\mathbb{C}^{N_{t}\times M}$ has
elements $[H_{t,q}]_{(n,m)}$, and its squared Frobenius norm is given
by $\|\mathbf{H}_{t,q}\|_{\mathrm{F}}^{2}=\sum_{n=1}^{N_{t}}\sum_{m=1}^{M}\left|[H_{t,q}]_{(n,m)}\right|^{2}$.
Under Rayleigh fading assumptions, the complex channel coefficients
$[H_{t,q}]_{(n,m)}\in\mathbb{C}$ for subcarrier $m\in\{1,\ldots,M\}$
and antenna $n\in\{1,\ldots,N_{\mathrm{t}}\}$ are assumed to be independent
and identically distributed (i.i.d.) complex Gaussian random variables.

In practical wireless systems, each AP operates in a unique environment,
leading to variations in channel propagation due to obstacles, building
materials, multipath effects, and antenna radiation patterns. Therefore,
the RSS measurement is modeled by a conditional path loss model that
accounts for the distinct propagation characteristics associated with
different APs and environments $s_{t,q}=\beta_{q}^{(k)}-\alpha_{q}^{(k)}\log_{10}d(\mathbf{x}_{t},\mathbf{o}_{q})+\xi_{q}^{(k)}$,
where $k\in\{0,1\}$ indicates the propagation mode (LOS for $k=0$
and NLOS for $k=1$). The term $\xi_{q}^{(k)}\sim\mathcal{N}(0,\sigma_{s,q,k}^{2})$ is to model the
randomness due to multipath fading, body shadowing, and
antenna pattern.

\subsubsection{Angle Model with LOS/NLOS Discrimination}

Consider the multipath propagation between the $q$-th AP and the
mobile user at location $\mathbf{x}_{t}$. The frequency-domain CSI
$\mathbf{H}_{t,q}\in\mathbb{C}^{N_{\mathrm{t}}\times M}$ is observed
over $M$ OFDM subcarriers and $N_{\mathrm{t}}$ transmit antennas.
The spatial covariance matrix is formulated as $\mathbf{R}_{t,q}=\frac{1}{M}\mathbf{H}_{t,q}\mathbf{H}_{t,q}^{\mathrm{H}}\in\mathbb{C}^{N_{t}\times N_{t}}$,
capturing the second-order statistics of the multipath channel. Through
eigendecomposition, we obtain $\hat{\mathbf{R}}_{t,q}=\tilde{\mathbf{u}}_{t,q}^{(1)}\lambda_{1}(\tilde{\mathbf{u}}_{t,q}^{(1)})^{\mathrm{H}}$,
where $\tilde{\mathbf{u}}_{t,q}^{(1)}$ is the orthonormal eigenvector
corresponding to the maximum eigenvalue $\lambda_{1}$. The noise
subspace matrix $\mathbf{U}_{t,q}$ is then constructed as $\mathbf{U}_{t,q}=\left[\tilde{\mathbf{u}}_{t,q}^{(2)},\tilde{\mathbf{u}}_{t,q}^{(3)},\dots,\tilde{\mathbf{u}}_{t,q}^{(N_{t})}\right]\in\mathbb{C}^{N_{t}\times(N_{t}-1)}$,
spanning the orthogonal complement of the signal subspace associated
with the largest eigenvalue. Based on MUSIC algorithm, the AoD of
the dominant path in the $q$-th AP coordinate system between the
$q$-th AP and mobile user at location $\mathbf{x}_{t}$ is estimated
by: 
\begin{equation}
\hat{\varphi}_{t,q}=\underset{\varphi\in(-\pi/2,\pi/2)}{\mathrm{argmax}}\frac{1}{\mathbf{a}^{\mathrm{H}}(\theta)\mathbf{U}_{t,q}\mathbf{U}_{t,q}^{\mathrm{H}}\mathbf{a}(\varphi)},\label{eq:H-theta}
\end{equation}
where $\mathbf{a}(\theta)$ is the steering vector given by (\ref{eq:steering}).
Thus, the AoD of the dominant path between the $q$-th AP and mobile
user at location $\mathbf{x}_{t}$ is estimated as $\hat{\theta}_{t,q}=\hat{\varphi}_{t,q}+\phi_{q}$.

Assuming the AoD distribution follows a conditional LOS/NLOS Gaussian
distribution $\hat{\theta}_{t,q}\sim\mathcal{N}\left(\phi(\mathbf{x}_{t},\mathbf{o}_{q}),\sigma_{\theta,k}^{2}\right),\quad(\mathbf{o}_{q},\mathbf{x}_{t})\in\mathcal{D}_{k}$,
where $\phi(\mathbf{x}_{t},\mathbf{o}_{q})$ defines the geometric
azimuth angle between the user location $\mathbf{x}_{t}$ and AP location
$\mathbf{o}_{q}$.The variance $\sigma_{\theta,k}^{2}$ quantifies
angular spread, which exhibits distinct behaviors in LOS ($k=0$)
and NLOS ($k=1$) regimes. The NLOS variance $\sigma_{\theta,1}^{2}$
typically exceeds the LOS variance $\sigma_{\theta,0}^{2}$ due to
multipath scattering in NLOS conditions.

\subsubsection{Delay Model with LOS/NLOS Discrimination}

Consider the channel $\mathbf{H}_{t,q}$ between the mobile user and
the $q$-th AP at time slot $t$. The variance features are extracted
as: 
\begin{equation}
\nu_{t,q}=10\log_{10}\left(\text{Var}\left(\left|\frac{\mathbf{H}_{t,q}}{\|\mathbf{H}_{t,q}\|_{2}}\right|\right)\right),\label{eq:H-nu}
\end{equation}
which quantifies multipath richness.

\subsection{An HMM Formulation}

Given the CSI $\mathbf{H}_{t,q}$, which can be separated as $\mathbf{y}_{t,q}=[s_{t,q},\theta_{t,q},\nu_{t,q}]^{\mathrm{T}}$
using (\ref{eq:H-s}), (\ref{eq:H-theta}) and (\ref{eq:H-nu}), where
$s_{t,q}$ represents the power, $\theta_{t,q}$ is the angle, and
$\nu_{t,q}$ is the delay components. According to the power-angle-delay
probability in Sec. \ref{subsec:PDFmodel}, we have the following
probability density function: 
\begin{align}
p(\mathbf{y}_{t,q}|\mathbf{x}_{t})=\begin{cases}
\mathcal{N}(\bm{\mu}^{(0)},\bm{\Sigma}^{(0)}), & (\mathbf{o}_{q},\mathbf{x}_{t})\in\mathcal{D}_{0},\\
\mathcal{N}(\bm{\mu}^{(1)},\bm{\Sigma}^{(1)}), & (\mathbf{o}_{q},\mathbf{x}_{t})\in\mathcal{D}_{1}.
\end{cases}\label{eq:prob-y}
\end{align}
where $\bm{\mu}^{(k)}(\mathbf{x}_{t})$ is the mean vector for the
$k$-th region (LOS or NLOS), defined as $\bm{\mu}^{(k)}(\mathbf{x}_{t})=[\beta_{q}^{(k)}+\alpha_{q}^{(k)}\log_{10}d(\mathbf{o}_{q},\mathbf{x}_{t}),\phi(\mathbf{x}_{t},\mathbf{o}_{q}),b_{k}+as_{t,q}]$,
and the covariance matrix $\bm{\Sigma}^{(k)}=\mathrm{Diag}(\sigma_{s,q,k}^{2},\sigma_{\theta,k}^{2},\sigma_{\nu}^{2})$.

Let $\mathcal{X}_{t}=(\mathbf{x}_{1},\ldots,\mathbf{x}_{t})$ represent
the trajectory of the user, and $\mathcal{H}_{t}=(\tilde{\mathbf{H}}_{1},\tilde{\mathbf{H}}_{2},\ldots,\tilde{\mathbf{H}}_{t})$
represent the accumulated channel measurements. Here, $\tilde{\mathbf{H}}_{t}=\{\mathbf{H}_{t,1},\mathbf{H}_{t,2},\dots,\mathbf{H}_{t,Q}\}$
denotes the set of all AP-side channels across subcarriers at time
slot $t$. The goal is to recover the complete trajectory $\mathcal{X}_{T}$
from the measurement history $\mathcal{H}_{T}$ using Bayesian inference.

We adopt the Gauss-Markov model for user mobility dynamics $\mathbf{x}_{t}$.
Let $\delta$ denote the slot duration. The movement at the $t$-th
time slot is modeled as: 
\begin{equation}
\mathbf{x}_{t}-\mathbf{x}_{t-1}=\gamma(\mathbf{x}_{t-1}-\mathbf{x}_{t-2})+(1-\gamma)\delta\bar{\mathbf{v}}+\sqrt{1-\gamma^{2}}\delta\bm{\epsilon},\label{eq:mobility}
\end{equation}
where the temporal spacing between consecutive locations $\mathbf{x}_{t-1}$
and $\mathbf{x}_{t}$ is denoted as $\delta_{t}$ seconds, the velocity
$\left(\mathbf{x}_{t}-\mathbf{x}_{t-1}\right)/\delta$ at time slot
$t$ depends on the velocity from the previous time slot, following
an auto-regressive model with parameter $0<\gamma\leq1$ and randomness
$\bm{\epsilon}\sim\mathcal{N}(\bm{0},\sigma_{v}^{2}\mathbf{I})$.
This captures the fact that acceleration is bounded in practice. The
parameter $\bar{\mathbf{v}}$ models the average velocity. A higher
$\gamma$ value indicates stronger correlation between consecutive
velocities, resulting in smoother movement. When $\gamma=1$, the
mobile user maintains a constant velocity.

The joint probability distribution decomposes through recursive application
of Bayes' theorem, yields $p(\mathcal{H}_{T},\mathcal{X}_{T})=\prod_{t=1}^{T}p(\tilde{\mathbf{H}}_{t}|\mathbf{x}_{t})\prod_{t=3}^{T}p(\mathbf{x}_{t}|\mathbf{x}_{t-1},\mathbf{x}_{t-2})$,
where $p(\tilde{\mathbf{H}}_{t}|\mathbf{x}_{t})$ is given by (\ref{eq:prob-y})
and $p(\mathbf{x}_{t}|\mathbf{x}_{t-1},\mathbf{x}_{t-2})$ is given
by (\ref{eq:mobility}).

Specifically, we consider the LOS/NLOS conditional spatial constrained,
and the joint probability $p(\mathcal{H}_{T},\mathcal{X}_{T})$ can
be written as:
\begin{align}
\tilde{p}(\mathcal{H}_{T},\mathcal{X}_{T}) & =\prod_{t=1}^{T}\prod_{q=1}^{Q}\prod_{k=0}^{1}p(\mathbf{y}_{t,q}|\mathbf{x}_{t})^{u_{t,q}^{(k)}}\prod_{t=3}^{T}p(\mathbf{x}_{t}|\mathbf{x}_{t-1},\mathbf{x}_{t-2}).\label{eq:tilde-p}
\end{align}
The logarithm of $\tilde{p}(\mathcal{H}_{T},\mathcal{X}_{T})$ can
be written as:
\begin{align*}
 & \mathcal{L}(\mathcal{X}_{T},\bm{\Theta}_{p},\bm{\Theta}_{m})=\sum_{t=1}^{T}\sum_{q=1}^{Q}\sum_{k=0}^{1}u_{t,q}^{(k)}\Bigg\{-\frac{1}{2}\Bigg[(\mathbf{y}_{t,q}-\bm{\mu}^{(k)}(\mathbf{x}_{t}))^{\mathrm{T}}\\
 & \qquad(\bm{\Sigma}^{(k)})^{-1}(\mathbf{y}_{t,q}-\bm{\mu}^{(k)}(\mathbf{x}_{t}))+\log|\bm{\Sigma}^{(k)}|+3\log(2\pi)\Bigg]\Bigg\}\\
 & \qquad-\sum_{t=3}^{T}\Bigg\{\log[2\pi\sigma_{\text{v}}\sqrt{1-\gamma^{2}}]\\
 & \qquad+\frac{\|\mathbf{x}_{t}-(1+\gamma)\mathbf{x}_{t-1}+\gamma\mathbf{x}_{t-2}-(1-\gamma)\delta\bar{\mathbf{v}}\|_{2}^{2}}{2(1-\gamma^{2})\delta^{2}\sigma_{\text{v}}^{2}}\Bigg\}
\end{align*}
where $\bm{\Theta}_{p}=\{\bm{\Phi}_{s},\bm{\Phi}_{\theta},\bm{\Phi}_{\nu},u_{t,q}^{(k)}\}$
represents the collection of propagation parameters. Here, $u_{t,q}^{(k)}$
is an auxiliary variable used to classify measurements into LOS or
NLOS regions. The last term represents a constraint on the spatial
structure.

Our goal is to maximize $\mathcal{L}(\mathcal{X}_{T},\bm{\Theta}_{\mathrm{p}},\bm{\Theta}_{\mathrm{m}})$
with the LOS/NLOS classification constraint: 
\begin{align}
\underset{\mathcal{X}_{T},\bm{\Theta}_{p},\bm{\Theta}_{m}}{\mathrm{maximize}} & \quad\mathcal{L}(\mathcal{X}_{T},\bm{\Theta}_{\mathrm{p}},\bm{\Theta}_{\mathrm{m}})\label{eq:P0}\\
\text{subject to} & \quad u_{t,q}^{(k)}\in\{0,1\},\;u_{t,q}^{(0)}+u_{t,q}^{(1)}=1.\nonumber 
\end{align}

\section{Algorithm}

To solve the joint trajectory recovery and parameter estimation problem
in equation (\ref{eq:P0}), we observe that, given $\mathcal{X}_{T}$,
the variables $\bm{\Theta}_{\text{{p}}}$ and $\bm{\Theta}_{\text{{m}}}$
are decoupled. This is because the first term in equation (\ref{eq:P0})
only depends on $\bm{\Theta}_{\text{{p}}}$, while the second term
only depends on $\bm{\Theta}_{\text{{m}}}$. Consequently, $\bm{\Theta}_{\text{{p}}}$
and $\bm{\Theta}_{\text{{m}}}$ can be solved through two parallel
subproblems derived from equation (\ref{eq:P0}), as follows: 
\begin{align*}
(\mathrm{P1}):\underset{\bm{\Theta}_{\text{{m}}}}{\mathrm{maximize}} & \;\;\sum_{t=3}^{T}\log p(\mathbf{x}_{t}|\mathbf{x}_{t-1},\mathbf{x}_{t-2};\bm{\Theta}_{\text{{m}}})\\
(\mathrm{P2}):\underset{\bm{\Theta}_{\text{{p}}}}{\mathrm{maximize}} & \;\;\sum_{t=1}^{T}\sum_{q=1}^{Q}\sum_{k=0}^{1}u_{t,q}^{(k)}\Bigg[-\frac{1}{2}\Bigg[(\mathbf{y}_{t,q}\\
 & \quad-\bm{\mu}^{(k)}(\mathbf{x}_{t}))^{\mathrm{T}}(\bm{\Sigma}^{(k)})^{-1}(\mathbf{y}_{t,q}-\bm{\mu}^{(k)}(\mathbf{x}_{t}))\\
 & \quad+\log|\bm{\Sigma}^{(k)}|+3\log(2\pi)\Bigg]\Bigg]\\
\text{subject to} & \quad u_{t,q}^{(k)}\in\{0,1\},\;u_{t,q}^{(0)}+u_{t,q}^{(1)}=1.
\end{align*}

On the other hand, given the variables $\hat{\bm{\Theta}}_{\text{{p}}}$
and $\hat{\bm{\Theta}}_{\text{{m}}}$ as the solutions to (P1) and
(P2), respectively, the trajectory $\mathcal{X}_{T}$ can be solved
by: 
\begin{align*}
(\mathrm{P3}):\underset{\mathcal{X}_{T}}{\mathrm{maximize}} & \;\;\mathcal{L}(\mathcal{X}_{T},\bm{\Theta}_{\mathrm{p}},\bm{\Theta}_{\mathrm{m}})
\end{align*}
This naturally leads to an alternating optimization strategy. In this
strategy, we solve for $\mathcal{X}_{T}$ from problem (P3), and then
for $\hat{\bm{\Theta}}_{\text{{p}}}$ and $\hat{\bm{\Theta}}_{\text{{m}}}$
from problems (P1) and (P2) iteratively. Since the corresponding iterations
never decrease the objective function in equation (\ref{eq:P0}),
which is bounded above, the iterations are guaranteed to converge.

The parameters $\{\bar{\mathbf{v}}, \sigma_{\text{v}}^{2}\}$ and $\{a_{1}^{(k)}, a_{2}^{(k)}\}$ in $\bm{\Theta}_{\mathrm{m}}$ are independent, and hence can be estimated separately. According to the mobility model in equation~(\ref{eq:mobility}), the probability density function (PDF) of the location $\mathbf{x}_{t}$ at time slot $t$ is uniquely determined by the Karush–Kuhn–Tucker (KKT) conditions.
Problem (P2) can be decomposed into three independent subproblems: (P2.1) with solution $\bm{\Phi}_{\nu}, u_{t,q}^{(k)}$, (P2.2) with solution $\bm{\Phi}_{s}$, and (P2.3) with solution $\bm{\Phi}_{\theta}$. 
To solve problem (P2.1), we adopt an iterative approach. Specifically, we first fix the binary variables $u_{t,q}^{(0)}$ and then optimize the parameters $b_{0}$, $b_{1}$, $a$, and $\sigma_{\nu}^{2}$. 

For each fixed $u_{t,q}^{(0)}$, this becomes a linear regression
problem. To solve for $b_{0}$, $b_{1}$, and $a$, we apply the least
squares method. The least squares problem for each class $k$ is:
\[
\underset{b_{0},b_{1},a}{\text{minimize}}\quad\sum_{t=1}^{T}\sum_{q=1}^{Q}u_{t,q}^{(k)}\left(\nu_{t,q}-b_{k}-as_{t,q}\right)^{2}
\]

This is a weighted least squares problem, and the solution is obtained
by solving for the parameters $b_{0}$, $b_{1}$, and $a$. We define
the matrix $\mathbf{S}_{k}$ and the observation vector $\mathbf{c}_{k}$.
Each row of $\mathbf{S}_{k}$ corresponds to an input pair $(1,s_{t,1})$
and each entry of $\mathbf{c}_{k}$ corresponds to the associated
response $\nu_{t,1}$, for $t=1,\ldots,T$. The weighted least squares
problem can be rewritten as:
\[
\underset{b_{k},a}{\text{minimize}}\quad\sum_{t=1}^{T}u_{t,q}^{(k)}\left(\mathbf{y}_{t}-\mathbf{S}_{k}[b_{k},a]^{\mathrm{T}}\right)^{2}
\]

The least squares solution for the parameters $[b_{k},a]^{T}$ is
given by the normal equation:
\begin{equation}
[b_{k},a]^{T}=\left(\mathbf{S}_{k}^{T}\mathbf{W}_{k}\mathbf{S}_{k}\right)^{-1}\mathbf{S}_{k}^{T}\mathbf{W}_{k}\mathbf{c}_{k}\label{eq:esti-ba}
\end{equation}
where $\mathbf{W}_{k}$ is the diagonal weighting matrix $\mathbf{W}_{k}=\text{diag}(u_{1,1}^{(k)},u_{2,1}^{(k)},\dots,u_{T,1}^{(k)})$.

To estimate the residual variance $\sigma_{\nu}^{2}$ for each class
$k$, we use the following formula:

\begin{equation}
\sigma_{\nu}^{2}=\frac{1}{T}\sum_{t=1}^{T}\sum_{k=0}^{1}u_{t,q}^{(k)}\left(\nu_{t,q}-b_{k}-as_{t,q}\right)^{2}\label{eq:esti-sigv}
\end{equation}

Next, fix the parameters $b_{0}$, $b_{1}$, $a$, and $\sigma_{\nu}^{2}$,
and optimize the binary variables $u_{t,q}^{(0)}$. The binary variables
are updated using a classification method such as Maximum Likelihood
Estimation (MLE) or Expectation-Maximization (EM). Thus, the binary
indicator $u_{t,q}^{(k)}$ is updated by searching for max $\exp\left(-\frac{(\nu_{t,q}-b_{k}-as_{t,q})^{2}}{2\sigma_{\nu}^{2}}\right)$. 

Repeat the alternative updating until convergence, thus solving problem
(P2.1). In each iteration, the parameters $b_{0}$, $b_{1}$, $a$,
and $\sigma_{\nu}^{2}$ are updated by solving the least squares problem,
and the binary variables $u_{t,q}^{(k)}$ are updated using the classification
method. This alternating process continues until the algorithm converges,
i.e., the updates to the parameters and binary variables no longer
change significantly.

We solve problem (P2.2) using MLE by decomposing it into independent
linear regression problems for each AP and set $k$. To solve problem
(P2.3), we employ MLE by separately estimating the noise variances
$\sigma_{\theta,0}^{2}$ and $\sigma_{\theta,1}^{2}$ by partitioning
the data $\{(\theta_{t,q},\phi(\mathbf{x}_{t},\mathbf{o}_{q}))\}$
into subsets $\mathcal{D}_{0}$ and $\mathcal{D}_{1}$ using the indicator
$u_{t,q}^{(k)}$. Here, $\mathcal{D}_{0}$ corresponds to $\sigma_{\theta,0}^{2}$
and $\mathcal{D}_{1}$ corresponds to $\sigma_{\theta,1}^{2}$.

Problem (P3) follows the classical \ac{hmm} optimization form, with
the distinction that the current state depends on the previous two
states. It can be efficiently solved using a modified version of the
Viterbi algorithm with a globally optimal guarantee.

The overall algorithm is summarized as follows: First, the propagation
parameters $\bm{\Theta}_{\text{{p}}}$ and mobility parameters $\bm{\Theta}_{\text{{m}}}$
are initialized randomly, and then the alternating update of $\mathcal{X}_{T}$,
$\bm{\Theta}_{\text{{p}}}$, and $\bm{\Theta}_{\text{{m}}}$ is performed
iteratively until convergence.

\section{Numerical Experiments}

\label{sec:Experiments}

\begin{figure}
\begin{centering}
\includegraphics[width=1\columnwidth]{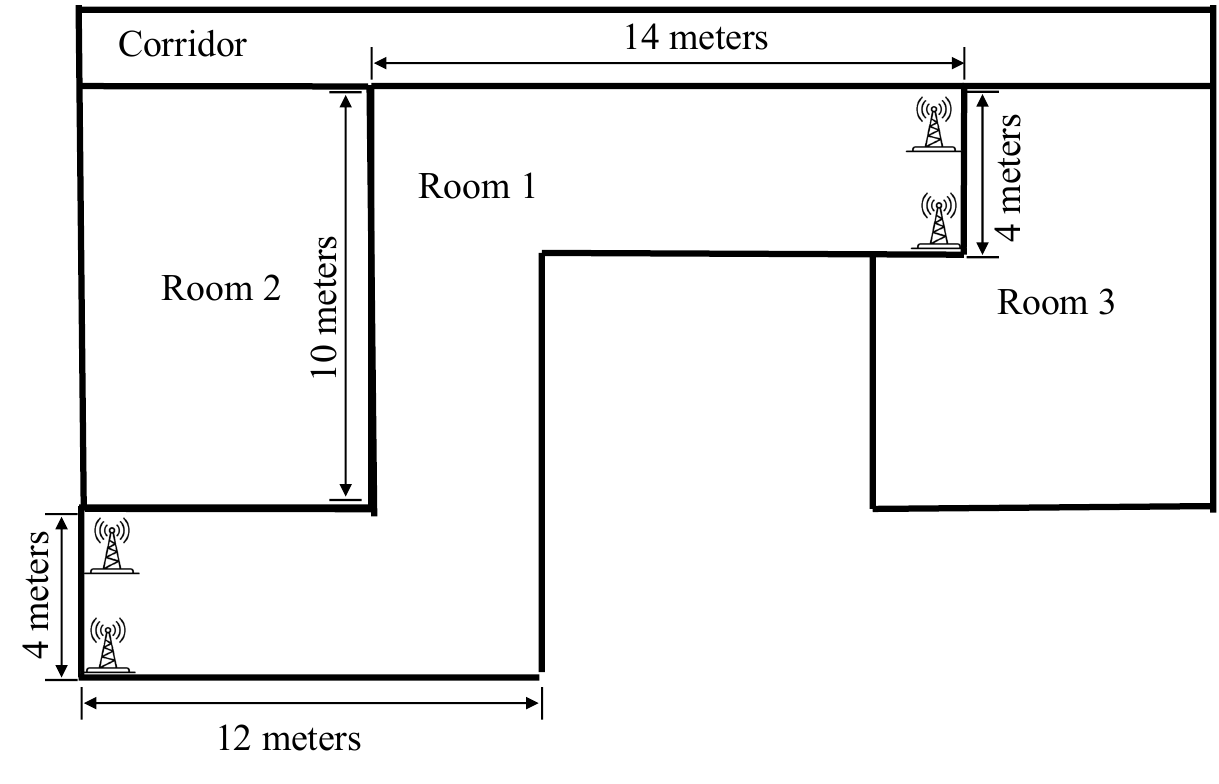}
\par\end{centering}
\centering{}\caption{Simulated indoor environment with LOS and NLOS regions. \label{fig:DatasetShow}}
\end{figure}

We utilized Wireless Insite$^{\circledR}$
to simulate an indoor environment with a 128 m$^{2}$ area. As illustrated
in Figure \ref{fig:DatasetShow}, four \acpl{ap} with a height
of 4 meters were manually deployed at the corners of the room. Each
\ac{ap} is equipped with an 8-antenna omnidirectional ULA array
and configured with $M=64$ subcarriers using a MIMO-OFDM model. The
antenna orientation of each \ac{ap} spans $180^{\circ}$, with
a transmit power of 0 dBm.  We set the antenna element spacing as $\Delta = 0.15\,\text{m}$, the carrier frequency as $f_{c} = 2.4\,\text{GHz}$, and the system bandwidth as $B = 20\,\text{MHz}$.

We recorded the CSI at receivers positioned at a height of 1.5 meters
along predefined trajectories with lengths of 167\,m, 160\,m, and
240\,m, corresponding to the training dataset, test dataset I, and
test dataset II, respectively. The training dataset measurements were
collected at walking speeds of 1, 1.5, 2, 2.5, 3, 3.5, and 4\,m/s,
with a sampling interval of $\delta=0.2$\,s. The trajectory in the
test dataset was generated by a random walk, with CSI measurements
corrupted by zero-mean Gaussian noise of variance 0.2 and 0.4, respectively. We executed the proposed algorithm with a discretization resolution of $\tau = 0.1$

\subsection{Trajectory Recovery Performance}

The proposed algorithm achieves a LOS/NLOS classification accuracy of $98.5\%$.
The trajectory estimation accuracy is quantified through the average
localization error metric $E_{\mathrm{loc}}=\frac{1}{T}\sum_{t=1}^{T}\|\mathbf{x}_{t}-\hat{\mathbf{x}}_{t}\|_{2}$,
where $\mathbf{x}_{t}\in\mathbb{R}^{2}$ denotes the ground-truth
position at time $t$, and $\hat{\mathbf{x}}_{t}$ represents the
estimated coordinates from our algorithm. Comparative analysis employs
four baseline methods, including: 1) Weighted Centroid Localization
(WCL) \cite{PhoSon:J18} with position estimates $\hat{\mathbf{p}}_{t}=\sum_{q=1}^{Q}w_{t,q}\mathbf{o}_{q}$
where $w_{t,q}=10^{s_{t,q}/20}/\big[\sum_{l=1}^{Q}10^{s_{t,l}/20}\big]$,
2) POG-AMP \cite{wang2022cooperative} leverage periodic AoA measurements
to estimated the user location. 3) Channel Charting \cite{taner2025channel}
embeds CSI into real-world coordinates by introducing a bilateration
loss and a line-of-sight bounding-box loss. 4) Genius-aided Map-Assisted
(GMA) method that alternately updates mobility parameters and trajectory
under known propagation models.

As shown in Table \ref{tab:traj-performance}, the proposed method
achieves the best overall performance, with an average localization
error of 0.65 meters. This result significantly outperforms the power-based
WCL method \cite{MagGioKanYu:J18}, which yields an average error
of 3.56 meters, as well as the angle-based POG-AMP approach \cite{wang2022cooperative}
and the Channel Charting method \cite{taner2025channel}, which report
errors of 1.67 and 1.65 meters, respectively. Among these benchmarks,
WCL performs poorly across all scenarios, especially under NLOS conditions
where its error reaches 4.74 meters. POG-AMP shows strong performance
in the Double LOS region with a low error of 1.25 meters, but suffers
in NLOS environments where multipath-induced angular deviations severely
degrade its accuracy, leading to a high error of 4.45 meters. Channel
Charting demonstrates relatively stable performance across different
LOS conditions, yet its lack of explicit delay modeling limits its
achievable precision.In contrast, the superior accuracy of our proposed
method stems from its integrated exploitation of power, angle, and
delay information, enabling a more holistic characterization of the
multipath propagation process. This joint modeling allows the system
to better distinguish between LOS and NLOS scenarios and enhances
localization robustness in complex environments.Although the proposed
method's average error is marginally higher than that of GMA (0.68
meters), which represents the theoretical upper bound under perfect
knowledge of propagation paths, the minimal difference of 0.03 meters
underscores the near-optimality of our approach in practical settings. 

\begin{table}[t]
	\centering{}\caption{Comparison of average localization error ($E_{loc}$) on the training
		dataset. \label{tab:traj-performance}}
	\begin{tabular}{>{\raggedright}V{\linewidth}|>{\centering}m{0.8cm}>{\centering}m{1.2cm}>{\centering}m{1.4cm}>{\centering}m{0.8cm}>{\centering}m{0.8cm}}
		\hline 
		& WCL \cite{PhoSon:J18} & POG-AMP\cite{wang2022cooperative} & Channel Charting \cite{taner2025channel} & Proposed & GMA\tabularnewline
		\hline 
		NLOS & 4.74 & 4.45 & 1.81 & 1.12 & 1.03\tabularnewline
		Single LOS & 3.81 & 2.94 & 1.79 & 0.77 & 0.74\tabularnewline
		Double LOS & 3.14 & 1.25 & 1.60 & 0.62 & 0.59\tabularnewline
		All & 3.56 & 1.67 & 1.65 & 0.65 & 0.68\tabularnewline
		\hline 
	\end{tabular}
\end{table}

\subsection{Localization Performance}

We evaluate the localization performance of the radio map using test
datasets I and II, which have not been used for the radio map construction.
A maximum-likelihood approach is used based on the conditional probability
function, and the estimated location $\hat{\mathbf{x}}$ given the
measurement vector $\mathbf{y}_{q}$ is given by maximizing the propagation
probability. 

We compare the localization performance with the unsupervised method
WCL \cite{PhoSon:J18}. For performance benchmarking, we also evaluate
three supervised localization approaches \ac{knn}, \ac{svm}\cite{SadSeb:J20},
and \ac{dnn}\cite{CheChu:J22}, which are trained using the training
set with location labels that were not available to the proposed scheme.
The parameters of baseline methods are determined and tuned using
a ten-fold cross validation. For \ac{knn}, the optimal number of
neighbors was found to be 8. A Gaussian kernel was used for \ac{svm}.
For \ac{dnn}, we adopt a three layer \ac{mlp} neural network
with 30 nodes in each layer to train the localization classifier.

\begin{table}[t]
\begin{centering}
\caption{Localization error {[}meters{]} on the test dataset I and II.\label{tab:Space-classification-performance}}
\par\end{centering}
\centering{}%
\begin{tabular}{>{\raggedright}m{2.7cm}|>{\centering}p{0.5cm}>{\centering}p{0.5cm}>{\centering}p{0.5cm}|>{\centering}p{0.4cm}>{\centering}p{1cm}}
\hline 
\multirow{1}{2.7cm}{Method} & \multicolumn{3}{c|}{Supervised} & \multicolumn{2}{c}{Unsupervised}\tabularnewline
 & KNN & SVM & DNN & WCL & Proposed\tabularnewline
\hline 
Test Dataset I & 0.81 & 0.84 & 0.85 & 3.82 & \textbf{ 0.80}\tabularnewline
Test Dataset II & 0.96 & 0.99 & 0.98 & 4.20 & \textbf{0.95}\tabularnewline
\hline 
\end{tabular}
\end{table}

Table~\ref{tab:Space-classification-performance} highlights three
key findings based on rigorous experimental validation. First, the
proposed unsupervised method significantly outperforms conventional
unsupervised approaches, achieving an impressive reduction in mean
localization error from 3.82 m to  0.80 m
compared to WCL. Second, while the performance of the proposed method
is comparable to that of supervised baselines, it enjoys the inherent
advantage of being label-free. This superiority arises from the method’s
ability to accurately model signal characteristics and incorporate
noise into the estimation process. As a result, it outperforms even
supervised methods on the noisy Test Dataset I. Third, although all
methods degrade on Test Dataset II due to increased signal noise,
the proposed method exhibits better robustness to noise, leading to
less performance degradation relative to other approaches. 

\section{Conclusion}

\label{sec:Conclusion}

This paper presents a novel unsupervised radio map construction method
based on HMM-driven trajectory recovery in MIMO networks, effectively
eliminating the need for location-labeled data calibration. By separately
modeling channel propagation under both LOS and NLOS conditions and
employing a Gaussian-Markov model to capture the spatial correlations
in user trajectories, the proposed approach accurately recovers both
propagation and mobility parameters through an alternating optimization
framework. Experimental results validate the effectiveness of the
method, achieving an average localization error of 0.65 meters, significantly
outperforming traditional baseline approaches. Furthermore, localization performance based on the constructed radio map exhibits a slight reduction in error compared to conventional supervised methods, which can be attributed to the superior noise-handling capability of the proposed approach.
Overall, this
work offers a robust and scalable solution to reduce calibration costs
and enhance localization accuracy in complex wireless environments. 

\bibliographystyle{IEEEtran}
\bibliography{Source/my_ref}

\end{document}